# Generating abbreviations using Google Books library

Valery D. Solovyev [a], Vladimir V. Bochkarev [a]

[a]*Kazan Federal University, Kremlyovskaya, 18, Kazan, 420008, Russia*

*E-mail address:* maki.solovyev@mail.ru, vbochkarev@mail.ru

**Abstract**

The article describes the original method of creating a dictionary of abbreviations based on the Google Books Ngram Corpus. The dictionary of abbreviations is designed for Russian, yet as its methodology is universal it can be applied to any language. The dictionary can be used to define the function of the period during text segmentation in various applied systems of text processing. The article describes difficulties encountered in the process of its construction as well as the ways to overcome them. A model of evaluating a probability of first and second type errors (extraction accuracy and fullness) is constructed. Certain statistical data for the use of abbreviations are provided.

### 1. Introduction

One of the first stages needed for numerous tasks of text processing is text segmentation into sentences and tokens – words and other combinations of symbols that henceforth should be appropriately processed as the single unit. The process of token allocation is called tokenization. The segmentation quality, including tokenization, impacts the effectiveness of solving various other tasks, such as morphological and syntactic analysis as well as extracting information from texts. At these stages of text processing, a hard problem [1] is the definition of the function of the period: if it is used to indicate the end of a sentence or for an abbreviation. Depending on that, segmentation can be different – the period makes a single token with the word abbreviation and develops into a separate token, if it is in the end of the sentence. Despite various undertaken research, no exact solution of this problem has been found. In complex cases, this problem is AI-complete, i.e. it requires, in general, the full knowledge of the world [1]. In this connection, most applied systems, such as Gate [2] and Uima (Apache UIMA project. http://uima.apache.org/ ), use simple heuristics for defining the function of the period as follows. If the period is followed by blank space, and then there is a word which starts with a capital letter, then the period is the end of the sentence. This is known as the period-space-capital pattern. Normally, it works well for simple fiction. Thus, in The Call of the Wild by Jack London, for example, this simple rule correctly identifies 98% sentence limits [3]. On the other hand, there can be cases when either a sentence ends with an abbreviation (then the period has a double function), or the period can be inside a sentence and is followed by a proper name. However, for more complex journalistic texts, such as the Wall Street Journal, results are worse – only 88.4% [1]. Therefore, the problem of identifying abbreviations and full words remains topical. In this article we describe a methodology of building an extensive dictionary of abbreviations. The research uses Russian language materials. For the sake of simplicity, in this work we will be limited by abbreviations composed of one word with the period. Abbreviations containing two and more periods can be analyzed as compositions of abbreviations with one period.

## 2. Problem statement and context of the work

A possible approach is the use of a dictionary and the search of full words therein. Abbreviations can also be identified in this dictionary as initial pieces of complete words. However, no matter how extensive a dictionary is, it will not be exhausting. Thus in [4] using the Google Books Ngram Corpus (http://books.google.com/ngrams/), we can see that modern English has more than one million words, while such fundamental dictionary as Webster Dictionary (2-nd edition, 2002) contains mere 348,000 words. Another approach is creating a possibly more complete dictionary of abbreviations. For English there are quite extensive databases on abbreviations, e.g. http://www.abbreviations.com containing 330,000 elements.

At the same time, currently, there is no extensive electronic dictionary of abbreviations for the Russian language that would be in common use. Available resources, such as (Russian Abbreviations Project, http://www.sokr.ru/about/), have a number of drawbacks. First of all, these are closed systems that do not give the user a file with the list of all abbreviations. Besides, normally, they do not differentiate between the abbreviations with and without the period. It is the first type which is crucial for the tasks of segmentation and tokenization. Our testing of the largest database of abbreviations in Russian (http://www.sokr.ru/about/) has revealed that it contains only a tiny fragment of actually used abbreviations with the period in texts. The similar drawbacks are also true for the databases of English abbreviations. Therefore, there is an urgent task of constructing a computer dictionary of abbreviations.

## 3. Methods and results

We propose a new method providing for the creation of an extensive dictionary of abbreviations (used with the period). For this purpose, we use the Google Books Ngram Corpus. It provides for the calculation of n-grams, even on the wide timescale. We use the data of the past 18 years. Bigrams made of a word (sequence of letters) followed by the period (periods in numbers are processed separately) are considered. Frequencies of the entire bigram and its left part before the period are correlated. In the Google Books Ngram Corpus, the period is always allocated into a separate token. If a bigram is an abbreviation, and its left part is not used without the period (which is our case for this study), then the compared frequencies coincide or almost coincide for the entire timescale. If in a bigram the period is just the end of the sentence and does not indicate the abbreviation, the compared frequencies vary significantly. The 90% threshold is used to compare the frequencies of the bigram and its left part, i.e. if in more than 90% of cases the word is used with the period, it is included into the dictionary of abbreviations. As a result of applying this method, there was produced a dictionary that contains approximately 9,000 abbreviations.

When the above approach is used in practice, several problems emerge. The Russian Google Books Ngram Corpus has quite a lot of misprints resulting from poor letter recognition by the recognition system used for scanned pages. This requires the secondary processing of the dictionary by hand. Furthermore, some abbreviations are occasionalisms, i.e. used extremely seldom, sometimes just in one book. Apparently, such abbreviations can be disregarded in systems of information extraction. Fig. 1 (a) shows the number of abbreviations used no more than in the given number of volumes. Applied systems may use various elimination thresholds by the number of volumes depending on the requirements of speed, fullness and other factors of text processing.

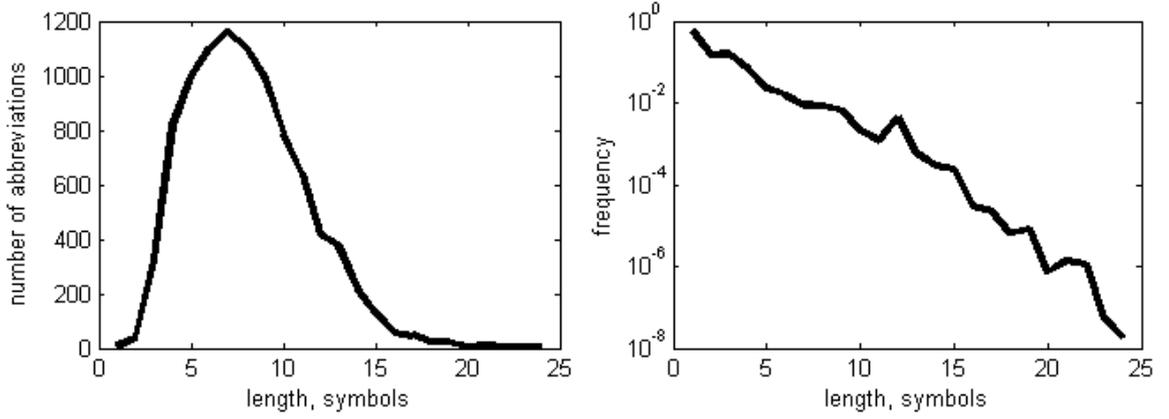

Figure 1. (a) Number of rare abbreviations used no more than in the given number of volumes (for 1990-2008); (b) Probability of usage of word forms with period

Some abbreviations were used in several volumes but published in one year or within a short period of time, which also points out that these abbreviations are not characteristic of the Russian language, in general. In order to have a numerical evaluation of the uniformity degree of the abbreviation usage for the entire time interval under consideration, an approach was applied that is based on the calculation of the median usage of an abbreviation. This also eliminates irrelevant abbreviations. The word form, which is knowingly an abbreviation, can be used with certain probability without the period. Besides, we should also consider errors caused by text recognition for the preparation of the Google Books Ngram Corpus. Sometimes, there were cases when the period was recognized as the comma and vice versa. If p is a probability of using a word form with the period, then the probability that from N cases in n cases it will be used with the period is described by the following binomial distribution:

$$P(n) = C_N^n p^n (1-p)^{N-n} \qquad (1)$$

Let us denote through $H_1$ a hypothesis that the selected word form is an abbreviation, and through $H_0$ - a hypothesis that we deal with a common word. We will use $p_1$ to denote a probability of using a word form with the period in the first case and $p_0$ - in the second. Then, the likelihood ratio shall take the following form:

$$L(n) = \frac{P(n|H_1)}{P(n|H_0)} = \left(\frac{p_1(1-p_0)}{p_0(1-p_1)}\right)^n \left(\frac{1-p_1}{1-p_0}\right)^N \qquad (2)$$

The rule of decision making is in comparing the likelihood ratio with a certain threshold *C*. Solving the equation $L(\eta)=C$, we can calculate the threshold value $\eta$. If $n>\eta$, a decision is made that the selected word form is an abbreviation. We can calculate the probability $\alpha(\eta)$ mistakenly taking a common word form for an abbreviation (probability of first type error):

$$\alpha(\eta) = P(n \geq \eta|H_0) = \sum_{n \geq \eta} C_N^n p_0^n (1-p_0)^{N-n} \qquad (3)$$

We also evaluate the probability $\beta(\eta)$ for missing an abbreviation (probability of second type error):

$$\beta(\eta) = P(n < \eta|H_1) = \sum_{n < \eta} C_N^n p_1^n (1-p_1)^{N-n} \qquad (3)$$

Thus, if we analyze the empirical data and find the probabilities $p_0$ and $p_1$, we will be able to construct a criterion for allocating abbreviations with the likelihood ratio needed for practice. In order to evaluate the probability $p_1$, a list of the most commonly used abbreviations was made. For each word form a share of the usage with the period was identified for each year from 1990 to 2008. The obtained annual values were later used to evaluate the median value for the share of the usage with the period for each word form. Afterwards, all word forms were sorted in the descending manner for this indicator. Let us note that for rather frequently used word forms the value of the median share of the usage with the period provides for a secure identification of the fact if the given word form is an abbreviation. Difficulties, primarily, emerge in case of relatively rare word forms. 300 word forms were manually selected from the beginning of the obtained list that are known and commonly used abbreviations in Russian. Abbreviations that have analogous writing of their complete forms were not included into the list, e.g. the word form "Инд" could mean the abbreviation from words "Indian", "India" or be the name of the river Indus. According to the obtained list of 300 abbreviations, values of probability $p_1$ were evaluated in different years. The share $1 - p_0$ of common word form usage without the period was evaluated. The results are presented in Fig. 1 (b) The mean value of $p_0$ for the period of 1998 to 2008 was 0.068, while the mean values of $p_1$ for the same period – 0.955. On the basis of the latter, in accordance with (3) we can calculate that the error level α=0.001 is achieved for the word form frequency usage of 40 and higher. In other words, among word forms used in the above period no less than 40 times, we will miss no more than 0.1% of abbreviations.

Various statistical regularities in the dictionary of abbreviations were investigated. Thus, Fig. 2 (a) shows the dependence of the number of abbreviations on the word length. It is obvious that in Russian most abbreviations fall on average sized words of 5-10 letters. Though most abbreviations are of average length, as it was expected, shorter abbreviations were most commonly used. Fig. 2 (b) indicates the data on the frequency of abbreviation usage as depending on the length. It is somewhat surprising that the dependence turned out to be practically linear (in logarithmic coordinates) which calls for further research to explain this phenomenon. Various statistical regularities in the dictionary of abbreviations were investigated.

Fig.3 shows the dynamics (for 1940-2008) for the usage of abbreviations in texts. Most frequently used abbreviations (dotted curve) were analyzed separately. These are the above mentioned 300 abbreviations. On average, they make up 75% of the total frequency of abbreviation usage. Qualitatively speaking, the graphs of short and full lists are similar. It seems interesting to find out why abbreviation usage soared in the mid-1990s.

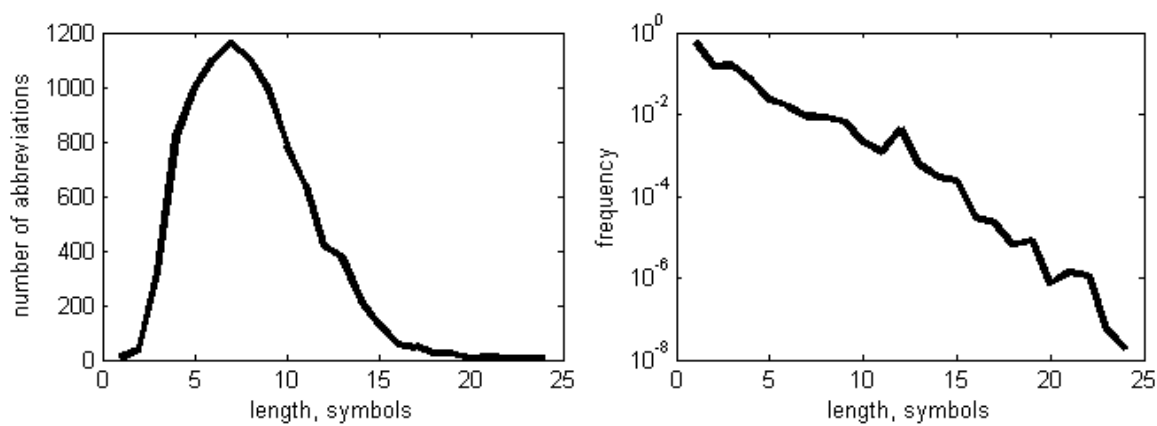

Figure 2. (a) Dependence of number of abbreviations on word length;
(b) Dependence of frequency of abbreviation usage on length

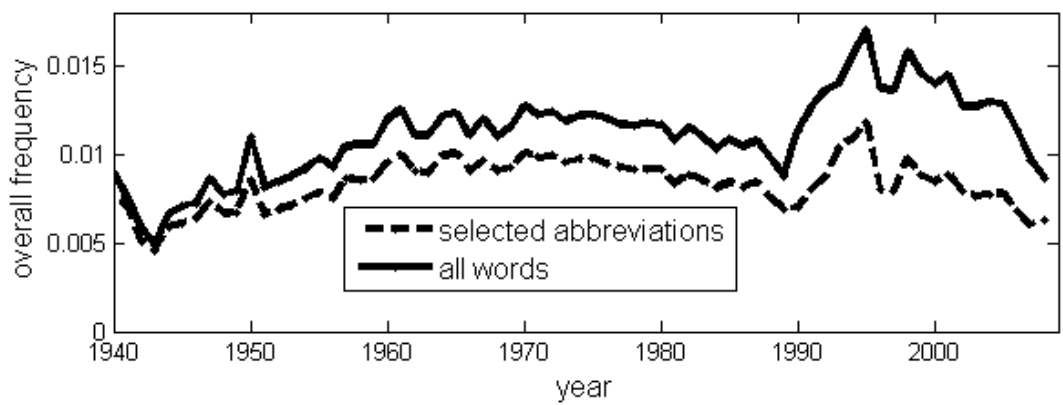

Figure 3. Sum frequency dynamics for usage of abbreviations

## 4. Conclusion

With the emergence of the super-extensive multilingual Google Books Ngram Corpus, there have appeared absolutely new opportunities for both theoretical and applied research. The article describes the method of creating the dictionary of abbreviations based on comparing the frequency of word usage in the Corpus with and without the period. The method is applied for constructing the dictionary of Russian abbreviations. This dictionary, when created, could be used in tokenization algorithms. Working with the Russian corpus, there emerge additional difficulties connected with a large number of errors caused by the recognition of characters. The methods of their correction are proposed. The obtained evaluations of first and second type error probability are the indicators of a rather high efficiency of the proposed method of abbreviation search. The statistical data on abbreviation usage are provided. It turned out that most abbreviations are made up of five to ten letters. However, in texts the shorter abbreviations the more frequently they are used, with dependencies having almost the exclusively linear character. The interesting data on abbreviation usage dynamics were obtained, yet their explanation needs further investigation.

**Acknowledgements**

The work was supported by the Russian Foundation of Basic Research, Grant № 12-06-00404-a.